\pdfoutput=1

\documentclass[11pt]{article}

\usepackage[preprint]{acl}

\usepackage{times}
\usepackage{latexsym}

\usepackage[T1]{fontenc}

\usepackage[utf8]{inputenc}

\usepackage{microtype}

\usepackage{inconsolata}

\usepackage{graphicx}

\usepackage{hyperref}
\usepackage{url}
\usepackage{adjustbox}
\usepackage{booktabs}
\usepackage{multirow}
\usepackage{algorithm}
\usepackage{algpseudocode}
\usepackage{verbatim}
\usepackage{philex}
\usepackage{caption}
\usepackage{subcaption}
\usepackage{listings} 
\usepackage{amsmath}

\title{Self-Taught Self-Correction for Small Language Models}

\author{
    \textbf{Viktor Moskvoretskii\textsuperscript{1,2}},
    \textbf{Chris Biemann\textsuperscript{3}},
    \textbf{Irina Nikishina\textsuperscript{3}}
    \\
\\
 \textsuperscript{1}Skoltech,
 \textsuperscript{2}HSE University,
 \textsuperscript{3}University of Hamburg
\\
 \small{
   \textbf{Correspondence:} \href{mailto:vvmoskvoretskii@gmail.com}{vvmoskvoretskii@gmail.com}
 }
}

\begin{document}
\maketitle
\begin{abstract}
Although large language models (LLMs) have achieved remarkable performance across various tasks, they remain prone to errors. A key challenge is enabling them to self-correct. While prior research has relied on external tools or large proprietary models, this work explores self-correction in small language models (SLMs) through iterative fine-tuning using solely self-generated data.  
We introduce the Self-Taught Self-Correction (STaSC) algorithm, which incorporates multiple algorithmic design choices. Experimental results on a question-answering task demonstrate that STaSC effectively learns self-correction, leading to significant performance improvements. Our analysis further provides insights into the mechanisms of self-correction and the impact of different design choices on learning dynamics and overall performance.  
To support future research, we release our user-friendly codebase and lightweight models.  

\end{abstract}

\section{Introduction}

Recent advanced LLM employ complex reasoning~\citep{guo2025deepseek} and meta-reasoning~\citep{xiang2025towards}, expanding their capabilities. 
However, even the most advanced models are prone to errors, including hallucinations~\citep{Huang_2025} and logical inconsistencies~\citep{ghosh2024logicalconsistencylargelanguage}, requiring external verification or human intervention. 
To address those problems, self-correction --- the ability to revise their own outputs --- has been evolved \citep{madaan2023selfrefineiterativerefinementselffeedback}.
The existing approaches mostly use zero-shot prompting~\citep{madaan2023selfrefineiterativerefinementselffeedback,shinn2024reflexion}, external evaluators for correction or feedback ~\citep{zhang2024small} or apply large proprietary models and focus specifically on mathematical tasks~\citep{kumar2024training}.

In this study, we focus on self-correction without external information or evaluators, ensuring inference efficiency while relying solely on the model’s intrinsic knowledge. We investigate the self-correction capabilities of small language models (SLMs) by applying iterative fine-tuning on self-generated data, allowing models to improve their outputs without external supervision.

We introduce the Self-Taught Self-Correction (STaSC) algorithm, which trains models to self-correct using self-generated trajectories, adapting the core idea of STaR~\citep{zelikman2022starbootstrappingreasoningreasoning}. STaSC extends and unifies approaches that iteratively train with self-generated trajectories, including the Self-Correction (SC) algorithm~\citep{welleck2022generating}, by incorporating flexible design choices. Unlike prior methods, STaSC provides control over initial answer exploration, correction filtering, and iterative fine-tuning, encompassing SC as a special case and demonstrating how different algorithmic choices impact self-correction performance.

\begin{figure*}[ht!]
    \centering
    \includegraphics[width=0.9\linewidth]{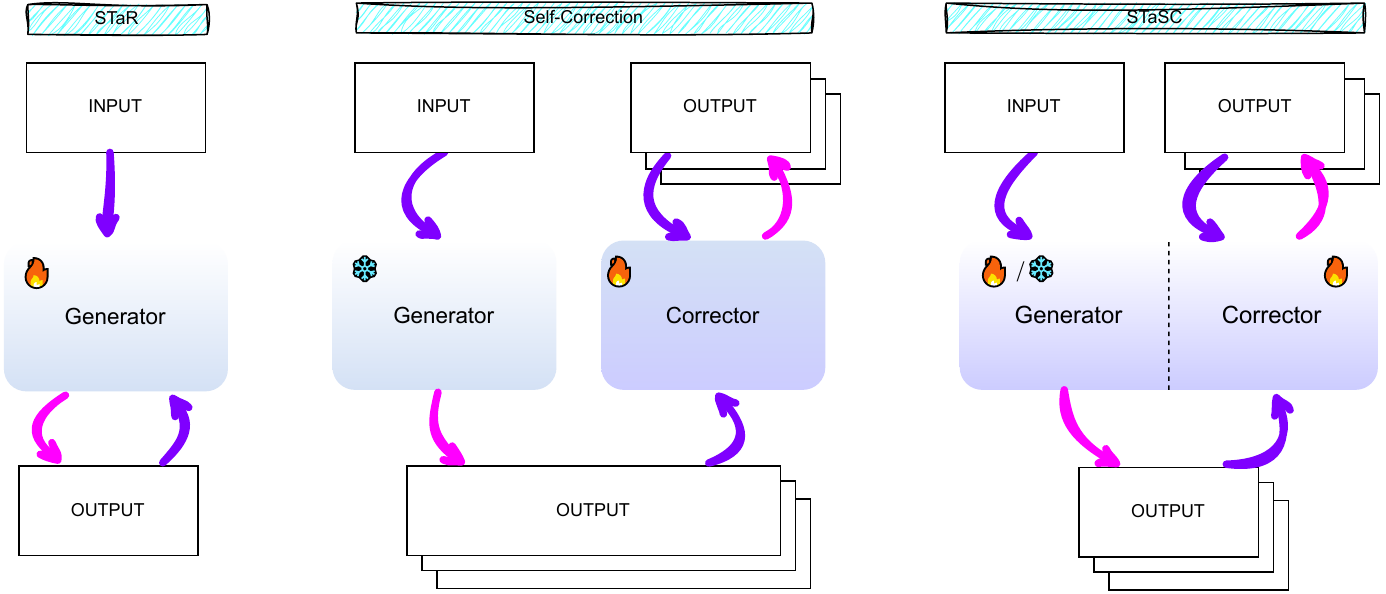}
    \caption{Illustration of the self-improvement method \textbf{STaR} (left) \cite{zelikman2022starbootstrappingreasoningreasoning}, self-correction method \textbf{SC} (center) \cite{welleck2022generating}, and our method, STaSC (right). STaCS offers flexible control over initial answer exploration,
correction filtering, and iterative fine-tuning. It is inspired by STaR and effectively encompasses SC as a special case. SC and STaSC allow several initial answers and corrections. The dotted line in the STaSC denotes two possible setups: fine-tuning the model and generating from it at the next iteration (Evolving Fine-Tuning) and keeping the Generator frozen and fine-tuning the Corrector model only (Fixed Fine-Tuning).}
    \label{fig:difference}
\end{figure*}

Our results on the \textit{Natural Questions} dataset \citep{kwiatkowski-etal-2019-natural} show that SLMs can learn to self-correct using self-synthesized data, while also improving their initial answer quality despite being trained solely for corrections. 
We release easy-to-use and adaptable code for self-correction algorithms at {\url{https://github.com/VityaVitalich/STASC}}.


The contributions of the paper are as follows:
\begin{itemize}
    \item We propose the Self-Taught Self-Correction (STaSC) algorithm, unifying and extending existing self-correction methods trained with self-generated trajectories.
    \item We conduct extensive experiments on a purely Natural Language Processing (NLP) task --- Question Answering --- using an open-source SLMs, demonstrating their ability to learn self-correction with self-synthesized data.
    \item We release open-source, easily adaptable code for self-correction, along with efficient SLMs with fewer than 4B parameters, making self-correction practical and accessible.
\end{itemize}

\section{Self-Taught Self-Correction}

In this section, we introduce the Self-Taught Self-Correction (STaSC) algorithm, an adaptation of STaR~\citep{zelikman2022starbootstrappingreasoningreasoning} for self-correction through iterative fine-tuning on self-synthesized data. STaSC unifies and extends various self-correction approaches using self-synthesized data, including the Self-Correction (SC) algorithm~\citep{welleck2022generating}.

\subsection{Foundations and Enhancements}

Figure \ref{fig:difference} presents three algorithms highlighting their similarities and dissimilarities. 
The left part describes the original \textbf{STaR} algorithm which focuses on generating reasoning paths, filtering correct ones, and fine-tuning them accordingly.

The central part of the Figure \ref{fig:difference} demonstrates the \textbf{Self-Correction (SC)} algorithm which fine-tunes the corrector model on corrections leading to improvement, keeping the initial Generator model fixed.
The right part refers to \textbf{STaSC}, adapting the idea of STaR of iteratively refining model outputs by sampling an initial answer, generating corrections, filtering correct trajectories, and fine-tuning on the successful revisions. This process is repeated over multiple iterations, progressively enhancing the model’s accuracy.

The key extension of our algorithm over previous baselines is the incorporation of flexible algorithmic choices. Initial answers can be drawn from either a frozen model or the previous iteration, corrections can be filtered based on strict improvements or by allowing non-worsening revisions, and fine-tuning can be performed from either the initial model or the latest iteration.

\subsection{Formal Definition and Algorithmic Choices}

Below, we outline the Algorithm~\ref{alg:stasc} steps and key design choices, each of which influences the self-correction process. For each choice, we define an abbreviation that will be used to denote specific algorithm configurations, such as STaSC\textsubscript{FIF}, where subscripts specify the selected options.

\subsubsection{Requirements and Notation}

The algorithm begins with an initial language model state, \( M_0 \), and an initial dataset, \( D_0 \), consisting of input-output pairs \((x, y)\). Additionally, we define a number of improvement iterations \( N \), the number of sampled initial generations \( N_{\text{init}} \), the number of sampled corrections \( N_{\text{corr}} \), and a reward function \( r \), which evaluates the quality of model-generated outputs.

\begin{algorithm*}[t!]
\caption{Self-Taught Self-Correction (STaSC)}
\label{alg:stasc}
\begin{algorithmic}[1]
\Require Initial model \( M_0 \), dataset \( D_0 \), number of iterations \( N \), 
initial samples \( N_{\text{init}} \), correction samples \( N_{\text{corr}} \), reward function \( r \)

\For{$n = 1$ to $N$}

    \State \textbf{Step 1: Sample Initial Answers}
    \State \(\hat{Y}^1_i = \{\hat{y}_{ij}^1\}_{j=1}^{N_{\text{init}}} \sim M(x_i)\),  \quad \( \forall x_i \in D_0 \)
    \State \quad \textbf{Option 1:} \( M = M_{n-1} \) \quad (Evolving Initialization)
    \State \quad \textbf{Option 2:} \( M = M_0 \) \quad (Fixed Initialization)

    \State \textbf{Step 2: Sample Corrections}
    \State \(\hat{Y}^2_i = \{\hat{y}_{ijk}^2\}_{k=1}^{N_{\text{corr}}} \sim M_{n-1}(x_i, \hat{y}_{ij}^1) \), \quad \( \forall \hat{y}_{ij}^1 \in \hat{Y}^1_i \)

    \State \textbf{Step 3: Filter Corrections}
    \State \( D_n^+ = \{(x_i, \hat{y}_{ij}^1, \hat{y}_{ijk}^2) \;|\; r(\hat{y}_{ijk}^2) > r(\hat{y}_{ij}^1) \} \)
    \State \( D_n^{=} = \{(x_i, \hat{y}_{ij}^1, \hat{y}_{ijk}^2) \;|\; r(\hat{y}_{ijk}^2) = r(\hat{y}_{ij}^1) \geq t \} \)
    \State \quad \textbf{Option 1:} \( D_n = D_n^+ \)  \quad (Improving Filter)
    \State \quad \textbf{Option 2:} \( D_n = D_n^+ \cup D_n^{=} \)  \quad (Non-Decreasing Filter)

    \State \textbf{Step 4: Fine-Tuning}
    \State \( M_n = \text{train}(M, \{\hat{y}_{ijk}^2 \;|\; (x_i, \hat{y}_{ij}^1, \hat{y}_{ijk}^2) \in D_n\}) \)
    \State \quad \textbf{Option 1:} \( M = M_0 \) \quad (Fixed Fine-Tuning)
    \State \quad \textbf{Option 2:} \( M = M_{n-1} \) \quad (Evolvong Fine-Tuning)

\EndFor
\end{algorithmic}
\end{algorithm*}

\paragraph{Step 1: Sampling Initial Answers}
In the first step, we sample \( N_{\text{init}} \) initial answers \( \hat{y}^1 \) for each input \( x \) in the dataset \( D_0 \). The primary design choice here is the selection of the model \( M \) used for sampling:

\begin{itemize}
    \item \textbf{Fixed Initialization (STaSC\textsubscript{F**})} – Initial answers are sampled using the frozen model \( M_0 \), as in the SC algorithm~\citep{welleck2022generating}.
    \item \textbf{Evolving Initialization (STaSC\textsubscript{E**})} – Initial answers are sampled using the model from the previous iteration \( M_{n-1} \), as in the original STaR~\citep{zelikman2022starbootstrappingreasoningreasoning}.
\end{itemize}

Fixed Initialization ensures robustness to variations in self-improvement and dataset shifts, maintaining consistency across iterations. In contrast, Evolving Initialization allows for greater exploration, potentially leading to more diverse refinements and improved performance.

\paragraph{Step 2: Sample Corrections} At the second step, we sample \( N_{\text{corr}} \) corrections \( \hat{y}^2 \) for each output \( \hat{y}^1 \) in dataset \( D_0 \) using the model from the last iteration \( M_{n-1} \).

\paragraph{Step 3: Filtering Corrections}
In this step, we filter the corrections using the reward function \( r(\hat{y}^2) \) to construct the fine-tuning dataset \( D_n \). The key design choice here is the filtering criterion for selecting corrections.

\begin{itemize}
    \item \textbf{Improving Filter (STaSC\textsubscript{*I*})} – Selects only corrections that strictly improve the reward, ensuring \( r(\hat{y}^2) > r(\hat{y}^1) \), as used in STaSC and SC.
    \item \textbf{Non-Decreasing Filter (STaSC\textsubscript{*N*})} – Selects corrections that either strictly improve the reward or retain the initial answer if it was already correct, ensuring \( (r(\hat{y}^2) = r(\hat{y}^1)) \cap (r(\hat{y}^1) \geq t) \), as proposed in SCoRE~\citep{kumar2024training}. This allows the model to preserve already correct answers while still refining incorrect ones.
\end{itemize}

The Improving Filter enforces strict improvement for every input, ensuring that only progressively better outputs are used for fine-tuning. In contrast, the Non-Decreasing Filter permits stable answers to be retained when already correct, allowing for a more conservative refinement process.

\begin{table*}[t]
    \centering
    \begin{tabular}{llcc|cc}
    \toprule
     \multirow{2}{*}{$N_\text{init}$} & \multirow{2}{*}{$N_\text{corr}$} & \multicolumn{2}{c|}{Qwen-2.5-1.5B} & \multicolumn{2}{c}{Phi3-mini} \\
     & & $max\{r(\hat{Y}^1)\}$  & $max\{r(\hat{Y}^2)\}$ & $max\{r(\hat{Y}^1)\}$  & $max\{r(\hat{Y}^2)\}$ \\
    \midrule
          \multirow{3}{*}{1} &       1 &             - &             - & $0.320 \pm 0.005$ & $0.372 \pm 0.010$ \\
           &       3 & $0.248 \pm 0.011$ & $0.208 \pm 0.011$ & $0.326 \pm 0.009$ & $0.376 \pm 0.010$ \\
           &       5 & $0.230 \pm 0.011$ & $0.228 \pm 0.021$ & $0.352 \pm 0.016$ & $0.384 \pm 0.012$ \\
    
           \midrule
          \multirow{3}{*}{3} &       1 & $0.236 \pm 0.007$ & $0.232 \pm 0.018$ & $0.334 \pm 0.011$ & $0.362 \pm 0.024$ \\
           &       3 & $0.264 \pm 0.015$ & $0.238 \pm 0.018$ & $0.342 \pm 0.010$ & $0.372 \pm 0.012$ \\
           &       5 & $0.273 \pm 0.017$ & $0.236 \pm 0.019$ & $0.332 \pm 0.007$ & $0.384 \pm 0.013$ \\
           \midrule
          \multirow{3}{*}{5} &       1 & $0.273 \pm 0.012$ & $0.250 \pm 0.024$ & $0.334 \pm 0.008$ & $0.378 \pm 0.016$ \\
           &       3 & $0.295 \pm 0.019$ & $0.244 \pm 0.023$ & $0.336 \pm 0.021$ & $0.354 \pm 0.026$ \\
           &       5 & $0.300 \pm 0.020$ & $0.248 \pm 0.029$ & $0.350 \pm 0.012$ & $0.376 \pm 0.011$ \\

    \bottomrule
    \end{tabular}
    \caption{Maximum reward $r$ over iterations for initial answer $r(\hat{Y}^1)$ and for correction $r(\hat{Y}^2)$ for different number of samples and initial generations. Bold corresponds to the best performance.}
    \label{tab:sampling_scores1}

\end{table*}

\paragraph{Step 4: Fine-Tuning}
In this step, we fine-tune the model on the dataset \( D_n \) obtained from Step 3 to produce the improved model \( M_n \). The key design choice here is the selection of the model used for fine-tuning:

\begin{itemize}
    \item \textbf{Fixed Fine-Tuning (STaSC\textsubscript{**F}}) – Fine-tunes the initial model \( M_0 \), as done in the original STaR recipe, ensuring stability across iterations and reducing the risk of accumulating errors.
    \item \textbf{Evolving Fine-Tuning (STaSC\textsubscript{**E}}) – Fine-tunes the model from the previous iteration \( M_{n-1} \), as in SC, allowing the model to progressively improve and adapt across iterations.
\end{itemize}

Fixed Fine-Tuning maintains a stable learning process by always training from the original model, preventing drift. In contrast, Evolving Fine-Tuning enables iterative adaptation, potentially leading to greater long-term improvements but also introducing the risk of compounding errors.


\section{Experimental Setup}

In this section, we describe the main experimental procedure, including dataset selection, evaluation metrics, and implementation details. 

\paragraph{Dataset} 
We evaluate our algorithm on the QA task using the {Natural Questions} dataset~\citep{kwiatkowski-etal-2019-natural}, which consists of factual simple questions. We use a subset of 500 questions per train and test split, following previous studies~\citep{trivedi2022interleaving,jeong2024adaptive,moskvoretskii2025adaptive}, to ensure consistency and computational efficiency. 

\paragraph{Evaluation}
We use In-accuracy as the primary evaluation metric and reward function, which is standard for this task~\citep{trivedi2022interleaving,jeong2024adaptive}. It assesses whether the generated answer contains the reference answer, assigning \( r(\hat{y}) = 1 \) for correct responses and \( r(\hat{y}) = 0 \) for incorrect ones.  

All metrics are reported on the test set, which remains unseen during training. Additionally, we report \( \max\{r(\hat{Y}^1)\} \) and \( \max\{r(\hat{Y}^2)\} \), representing the highest reward obtained for initial answers and corrections across STaSC iterations, respectively.  
To ensure a fair real-world setup, the reward is not available to the model during inference. Unlike some prior studies~\citep{shinn2024reflexion}, where inference benefits from reward signals derived from ground-truth labels, our approach aligns with a fully unsupervised inference setting, ensuring a more realistic evaluation.

\paragraph{Implementation Details} 
We conduct experiments using Qwen-2.5-1.5B~\citep{qwen2025qwen25technicalreport} and Phi3-Mini~\citep{abdin2024phi3technicalreporthighly}, 
employing default generation parameters. The default setup for self-correction is 2-shot. STaSC fine-tuning is performed for 1 epoch with a batch size of 8 and a learning rate of \( 7 \times 10^{-6} \). Additional implementation details are provided in Appendix~\ref{sec:appendix_technical}.  


\begin{figure*}[t]
    \centering
    \includegraphics[width=\linewidth]{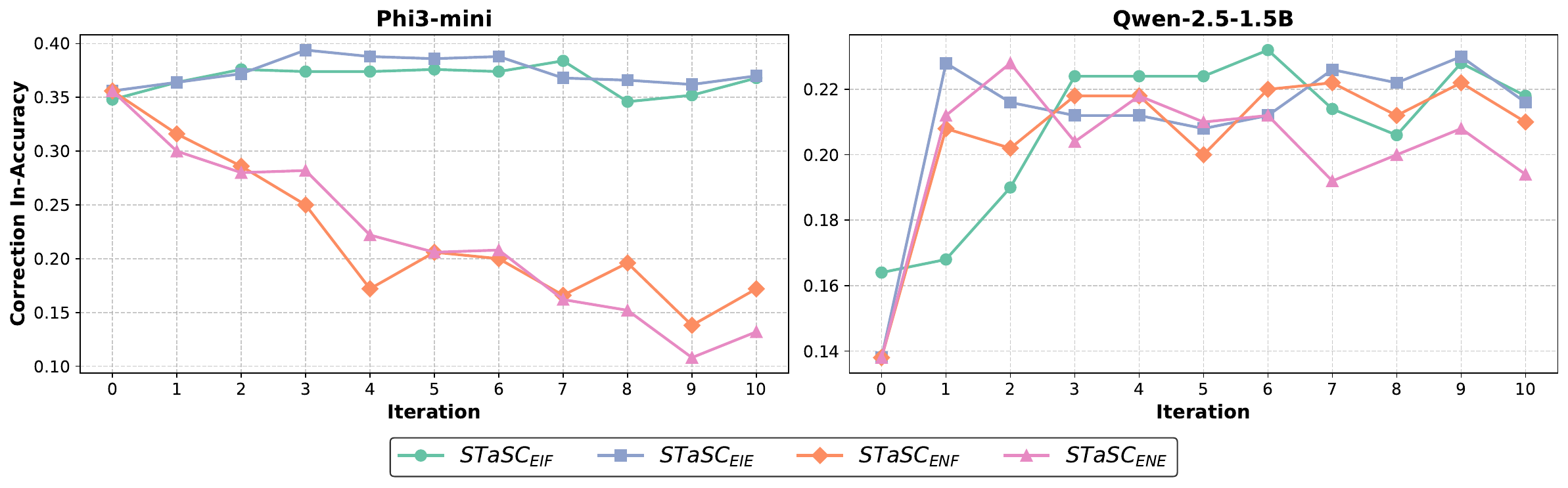}
    \caption{Correction In-accuracy for STaSC versions with Evolving Initialization for Phi3-mini and Qwen-2.5-1.5B.}
    \label{fig:evolving}
\end{figure*}  

\begin{figure*}[t]
    \centering
    \includegraphics[width=\linewidth]{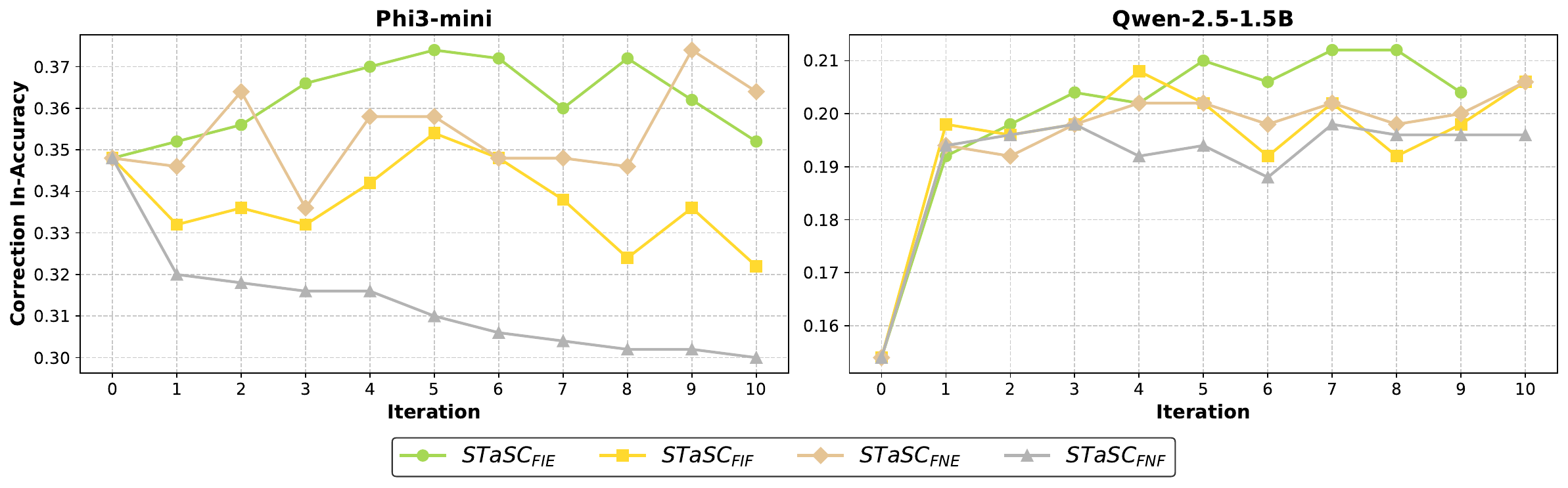}
    \caption{Correction In-accuracy for STaSC versions with Fixed Initialization for Phi3-mini and Qwen-2.5-1.5B.}
    \label{fig:stasc_fixed}
\end{figure*}

\begin{figure*}[t]
    \centering
    \includegraphics[width=\linewidth]{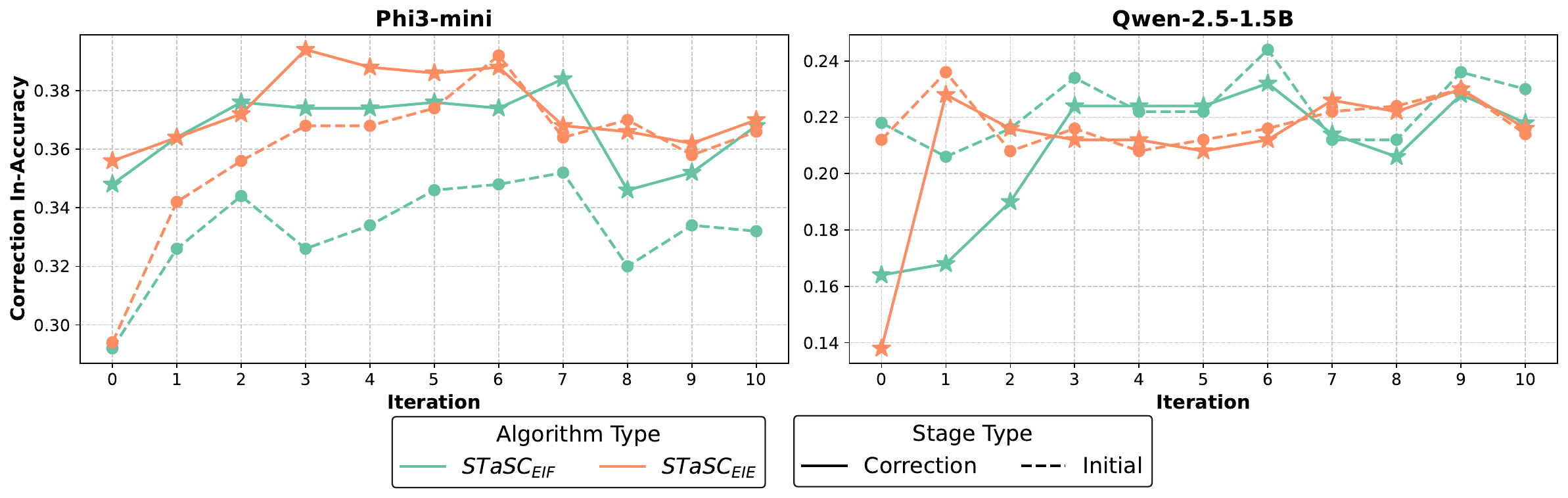}
    \caption{Correction and Initial Answer In-accuracy for best STaSC versions for Phi3-mini and Qwen-2.5-1.5B.}
    \label{fig:stasc_best}
\end{figure*}

\section{Results \& Discussion}
In this section, we provide the results and discuss them, inspecting the \textit{STaSC} algorithm design.

\subsection{Impact of $N_\text{init}$ and $N_\text{corr}$}

Firstly, we examine how the selection of parameters \( N_\text{init} \) and \( N_\text{corr} \) affects algorithm performance, exploring values of 1, 3, and 5 for both models. To encourage exploration, we use STaSC\textsubscript{EIF}, where initial answers are sampled from the previous iteration, only improving corrections are retained, and fine-tuning is performed from the base model to ensure stability.

As shown in Table~\ref{tab:sampling_scores1}, a greedy approach for \textit{Qwen-2.5-1.5B} fails to ensure convergence, as no improving corrections emerge in the first iteration. In contrast, increased exploration significantly enhances performance, likely due to the weaker alignment of the initial model. Additionally, we observe that exploring initial answers has a greater impact than exploring corrections.

However, this trend does not hold for \textit{Phi3-Mini}, where increasing the number of initial answers does not improve results, but greater correction exploration does. This discrepancy likely stems from differences in model capabilities. Phi3-Mini, being inherently stronger, benefits more from refining corrections, whereas \textit{Qwen-2.5-1.5B}, with lower initial competence, requires a broader search for initial answers to gain useful knowledge early on.

For subsequent experiments, we adopt the most stable and well-performing configurations:
\begin{itemize}
    \item \textit{Qwen-2.5-1.5B}: \( (N_\text{init}, N_\text{corr}) = (5, 5) \).
    \item \textit{Phi3-Mini}: \( (N_\text{init}, N_\text{corr}) = (1, 5) \).
\end{itemize}

\begin{table*}[t!]
    \centering
    \resizebox{\textwidth}{!}{%
    \begin{tabular}{ccccc|cc}
    \toprule
     Step 1  & Step 3  & Step 4  & \multicolumn{2}{c|}{Qwen-2.5-1.5B} & \multicolumn{2}{c}{Phi3-mini} \\
     Model & Filter & Model & {$max\{r(\hat{Y}^1)\}$} & {$max\{r(\hat{Y}^2)\}$} & {$max\{r(\hat{Y}^1)\}$} & {$max\{r(\hat{Y}^2)\}$} \\
    \midrule
         \multirow{4}{*}{$M_{0}$} & \multirow{2}{*}{Improving} & $M_{0}$ &  \( 0.212 \) & \( 0.208 \pm 0.014 \) & $0.294$ & $0.354 \pm 0.010$\\
       &  & $M_{n-1}$ &  \( 0.212 \) & \( 0.212 \pm 0.016 \) & $0.294$&  $0.374 \pm 0.009$\\
       \cmidrule{2-7}
       & \multirow{2}{*}{Non-Decreasing} & $M_{0}$ &  \( 0.212 \) & \( 0.198 \pm 0.012 \) & $0.294$ & $0.348 \pm 0.013$\\
       &  & $M_{n-1}$ &  \( 0.212 \) & \( 0.206 \pm 0.014 \) & $0.294$ & $0.374 \pm 0.010$\\

    \midrule

     \multirow{4}{*}{$M_{n-1}$} & \multirow{2}{*}{Improving} & $M_{0}$ &  \( \mathbf{0.244 \pm 0.011} \) & \( \mathbf{0.232 \pm 0.023} \) & \underline{$0.352 \pm 0.016$} & \underline{$0.384 \pm 0.12$}\\
    &  & $M_{n-1}$ &  \( 0.236 \pm 0.009 \) & \underline{\( 0.230 \pm 0.024 \)} & $\mathbf{0.392 \pm 0.024}$& $\mathbf{0.394 \pm 0.012}$ \\
       
        \cmidrule{2-7}
       & \multirow{2}{*}{Non-Decreasing} & $M_{0}$ &  \underline{\( 0.240 \pm 0.009 \)} & \( 0.222 \pm 0.023 \) & $0.316 \pm 0.056$ & $0.356 \pm 0.066$\\
       
       &  & $M_{n-1}$ &  \( 0.234 \pm 0.013 \) & \( 0.228 \pm 0.022 \) & $0.294 \pm 0.062$ & $0.356 \pm 0.074$\\

    \bottomrule
    \end{tabular}
    }
    \caption{Maximum reward $r$ over iterations for initial answer $r(\hat{Y}^1)$ and for correction $r(\hat{Y}^2)$ for different settings of STaSC Algorithm. Bold values correspond to the best performance, underlined represent second best.}
    \label{tab:main_stasc}

\end{table*}

\subsection{STaSC With Evolving Initialization}

Figure~\ref{fig:evolving} illustrates the dynamics of correction performance for STaSC variants with Evolving Initialization. Below, we discuss the observed effects on performance, highlighting key trends and their implications.

\paragraph{Effect of the Non-Decreasing Filter}  
A key observation is that STaSC with the Non-Decreasing Filter consistently degrades performance for Phi3-Mini. This is likely due to the difficulty in stabilizing training when fewer corrections are retained, increasing the risk of overfitting. Interestingly, Qwen does not exhibit the same decline, possibly due to a higher number of retained corrections, which augments the data and stabilizes training.

\paragraph{Impact of Filtering Selectivity}  
We further highlight the need to properly filter corrections when using Evolving Fine-Tuning, as all Phi3 settings with STaSC\textsubscript{**E} exhibit a negative correlation between the leniency of filtered trajectories and the gain in correction performance (\( r=-0.51, p < .001 \)). The more corrections were provided to the Corrector model, the worse it performed. This suggests that insufficiently selective filtering introduces noise, leading to overfitting and a decline in performance improvement. Notably, no other setting shows a significant correlation between the number of filtered corrections and performance.

\paragraph{Evolving Fine-Tuning Trends}  
Evolving Fine-Tuning slightly improves correction performance for Phi3-Mini. For Qwen, performance increases rapidly at first, then declines slightly before rising again in later iterations. This suggests that accumulated knowledge gains take effect only in the later stages, once a sufficient number of corrections and initial answers have been processed.

\paragraph{Key Takeaways}  
\textit{Evolving Initialization is most effective when combined with an Improving Filter and Evolving Fine-Tuning, as these either enhance correction performance or at least prevent degradation. Additionally, filtering selectivity is crucial—overly lenient filters introduce noise and cause overfitting.}

\subsection{STaSC With Fixed Initialization}

Next, we analyze the trends and implications of Fixed Initialization, which aligns with the algorithmic choice used in SC. Figure~\ref{fig:stasc_fixed} presents the correction performance for STaSC variants under this setting. 

\paragraph{Effect of the Non-Decreasing Filter}  
Unlike Evolving Initialization, the Non-Decreasing Filter does not lead to a general decline in correction performance for either Phi3 or Qwen, with an exception of STaSC\textsubscript{FNF},where performance degrades. 
For Phi3, correction performance remains largely unchanged throughout most iterations, with noticeable improvements emerging only in the final stages. This delayed progress suggests that Evolving Fine-Tuning gradually accumulates knowledge, but its effects become apparent only after sufficient corrections and initial answers have been processed. A similar trend was previously observed in STaSC\textsubscript{EIE}, showing that iterative fine-tuning plays a key role in long-term performance gains.

\paragraph{Importance of Evolving Fine-Tuning}  
We find that Evolving Fine-Tuning is crucial when using Fixed Initialization, particularly for Phi3 and, to a lesser extent, for Qwen. This is expected, as Evolving Fine-Tuning serves as the only source of exploration in this setting, driving the algorithm forward. In contrast, with Fixed Fine-Tuning, we observe a general decline in Phi3’s performance and stagnation after the first iteration for Qwen. This suggests that without sufficient exploration, relying solely on corrections from previous steps is insufficient for SLMs.

We also observe that when using both Fixed Initialization and Fixed Fine-Tuning applying the Non-Decreasing Filter further worsens the performance, underscoring the importance of filtering selectivity. In this setting, exploration is driven solely by corrections, making the filtering process critical. When corrections include too many noisy or uninformative trajectories due to improper filtering, the model struggles to improve, leading to significant performance degradation. This reinforces the need for a more controlled correction selection process to ensure meaningful updates during training.

\paragraph{Key Takeaways}  
\textit{Fixed Initialization reduces exploration, making the model more tolerant to the Non-Decreasing Filter and benefiting more from Evolving Fine-Tuning. In contrast, Fixed Fine-Tuning restricts exploration solely to corrections, increasing reliance on selective filtering to maintain performance.}

\begin{figure*}[ht!]
    \centering
    \includegraphics[width=0.95\linewidth]{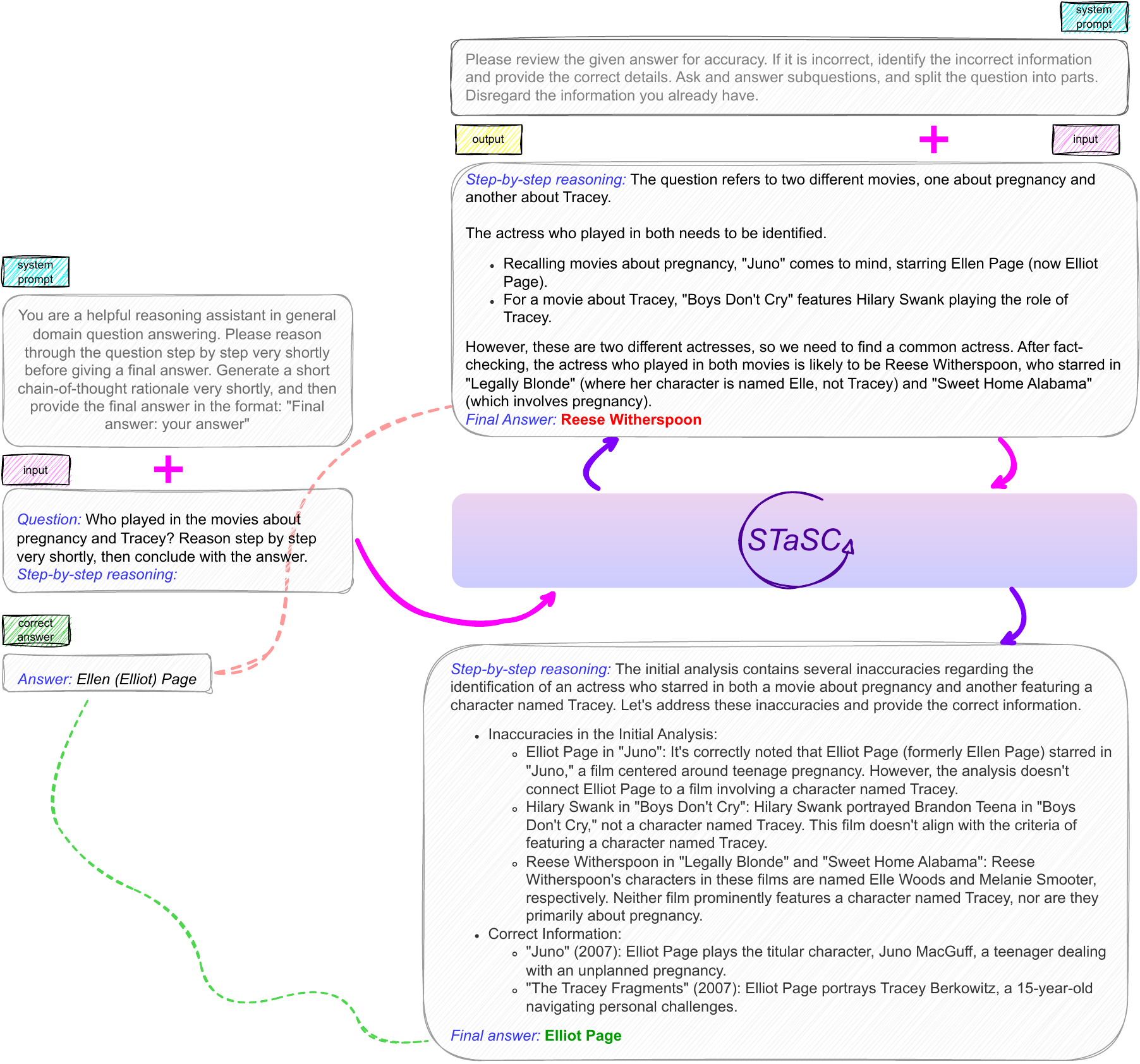}
    \caption{Example of the STaSC pipeline.}
    \label{fig:example}
\end{figure*}

\subsection{STaSC Impact on Initial Answers}
In this section, we identify the best-performing STaSC configurations and analyze their behavior in terms of initial answers and correction dynamics.

\paragraph{Selection of High-Performing Variants}  
Based on our previous analysis and Table~\ref{tab:main_stasc} we determine that STaSC\textsubscript{EIE} and STaSC\textsubscript{EIF} exhibit the strongest performance. These results highlight the crucial role of Evolving Initialization and the Improving Filter, while leaving the impact of Fine-Tuning strategies open for further investigation.

\paragraph{Performance Comparison}  
Figure~\ref{fig:stasc_best} illustrates the performance trends of initial answers and corrections for STaSC\textsubscript{EIE} and STaSC\textsubscript{EIF}.  

The effectiveness of Fine-Tuning strategies varies between Phi3 and Qwen. For Phi3, Evolving Fine-Tuning leads to a substantial increase in initial answer quality, surpassing Fixed Fine-Tuning, while yielding  moderate improvements in corrections. In contrast, for Qwen, Fixed Fine-Tuning results in superior performance for both initial answers and corrections.

\paragraph{Effect of Correction-Only Training}  
Interestingly, despite the model being trained exclusively on corrections, initial answer quality also improves. Since gradient updates are applied only to correction tokens, this suggests that learning corrections either enhances the model’s factual knowledge or improves its internal reasoning ability at the initial generation stage.

\paragraph{Alignment Between Initial Answers and Corrections}  
Evolving Fine-Tuning progressively reduces the gap between initial answers and corrections, eventually leading to their alignment. For Phi3, initial answers gradually improve until they reach the quality of corrections. For Qwen, the opposite trend is observed—corrections improve until they match the initial answers.  

This suggests that Evolving Fine-Tuning either helps internalize the correction process, leading to higher-quality initial answers as seen in Phi3, or stabilizes responses by preventing degradation, as observed in Qwen.

\section{Related Works}

Self-correction is a relatively new yet actively growing research domain. A systematic survey defines self-correction as a framework in which LLMs refine their responses using LLMs during inference, potentially incorporating external tools or knowledge~\citep{kamoi2024can}. A significant body of work in this direction focuses on leveraging external feedback from external knowledge and verification tools~\citep{jiang2023activeretrievalaugmentedgeneration,gou2024criticlargelanguagemodels,pan2023logic,xu2023instructscore}, as these approaches provide high-quality solution evaluation. However, in real-world applications, such external resources are often unavailable or computationally expensive. Moreover, relying on external verification does not pose a fundamentally challenging task for LLMs, limiting their ability to develop intrinsic reasoning and self-improvement capabilities.

An alternative approach is intrinsic self-correction, where the model refines its own outputs without relying on external critics. This can be implemented in a zero-shot setting, using the same model iteratively~\citep{madaan2023selfrefineiterativerefinementselffeedback}, or through external models trained on synthetic errors\citep{paul-etal-2024-refiner} or self-generated corrections\citep{welleck2022generating}. While these approaches have shown promise, they still rely on external critic models, making them closer to verification-based generation rather than true self-correction.

The only work exploring self-correction in its natural form is the SCoRE framework~\citep{kumar2024training}, which was the first to experiment with intrinsic self-correction within the same model and introduced a multi-turn RL training approach for this purpose. However, SCoRE lacks a formalized theoretical foundation and a deeper investigation of baseline algorithm adaptations. Additionally, it is limited by the use of large proprietary models, without open-sourcing the code or model weights, making it difficult to build upon for future research.

\vspace{-0.1cm}
\section{Conclusion}

In this study, we introduced the Self-Taught Self-Correction (STaSC) algorithm, which incorporates multiple algorithmic choices to enable genuine intrinsic self-correction without relying on external tools or large proprietary models. Inspired by STaR~\citep{zelikman2022starbootstrappingreasoningreasoning}, our approach trains exclusively on self-generated data.  
Experiments on a QA task with two small language models demonstrate that SLMs can learn to self-correct using their own generations and even improve initial answers, despite being trained solely for corrections. Furthermore, our analysis highlights key algorithmic insights, emphasizing the importance of filtering selectivity, initial answer exploration, and the potential of iterative fine-tuning.  
To support future research, we have open-sourced our code and lightweight models.

\section{Limitations}

\begin{itemize}
\item The selected SLMs, while effective, may have certain capacity limitations that could influence the extent of the self-correction process.  
\item Experiments were conducted with a single run, which, while sufficient for initial insights, may introduce some variability.  
\item The evaluation focuses on a Question Answering (QA) task, leaving open the opportunity to explore performance across other tasks and domains.  
\item The chosen hyperparameters, though reasonable, may not fully optimize the model's learning efficiency or overall performance.  
\item A more detailed analysis of the types and patterns of corrections could further enrich our understanding of the self-correction mechanism.  
\item The reward function, while practical, may not perfectly capture all nuances of desired behavior, presenting room for refinement in future work.  
\end{itemize}

\section{Ethical Considerations}

Our work enables small language models to self-correct using self-generated data. We employ advanced models like Qwen-2.5 and Phi-3, pre-trained on diverse datasets, including user-generated content. While efforts have been made to remove harmful or biased data, some biases may persist in outputs. This does not undermine our methods, which are designed to self-correct factual inaccuracies and are adaptable to other rigorously debiased models. Beyond inherent bias challenges, our work raises no additional ethical concerns.

\bibliography{custom}

\appendix

\section{Technical Details} \label{sec:appendix_technical}

We apply a weight decay of 0.1 and use the Adam optimizer with default betas, along with a cosine annealing scheduler. Training is performed using Fully Sharded Data Parallel on two A100 GPUs, with an estimated total compute of 80 GPU hours.
The dataset and Qwen come under Apache License, Phi3-mini under MIT. We are open-sourcing all our code under the Apache License. 

\paragraph{Initial answer sampling prompt}
\begin{quote}
You are a helpful reasoning assistant in general domain question answering. Please reason through the question step by step very shortly before giving a final answer.

Generate a short chain-of-thought rationale very shortly, and then provide the final answer.

{Step-by-step reasoning:} \\
{Final Answer:} \\

{Question:} \{question text\} \\

Reason step by step very shortly, then conclude with the answer.
\end{quote}

\paragraph{Correction sampling prompt}
\begin{quote}
You are a helpful reasoning assistant in general domain question answering.Your task is to correct the initial response if it is incorrect.

Below is the question and the initial answer. Generate a correction to the initial answer if it is incorrect. Disregard the information you already have, look for other options. Do not use the information that does not match your criteria. 

Step-by-step reasoning:

Final Answer:

Question: Which Portuguese soccer player has the most goals?

Step-by-step reasoning: Christiano Ronaldo has scored 900 goals in a career spanning 22 years. I cannot think of anyone else scoring more than him.

Final Answer: Christiano Ronaldo

Step-by-step reasoning: Yes, Christiano Ronaldo is the correct answer. Other details are unimportant.

Final Answer: Christiano Ronaldo

Question: Which actress played in both movies: one about preg-cy and the other about Tracey?

Step-by-step reasoning: 1. Identify movies about preg-cy and about Tracey. 2. Movies about preg-cy include 'Juno' and 'Knocked Up'. 3. Movie about Tracey is likely 'Tracy Chapman' or a documentary, but 'Tracy' could be a reference to 'Tracy Flick' in the movie 'Election'. 4. Considering well-known movies with these themes, actress Reese Witherspoon comes to mind as a possible match. 

Final Answer: Reese Witherspoon

Step-by-step reasoning: Let's break down the answer. 'Juno' is indeed a movie about preg-cy, but it starred Ellen Page (now Elliot Page), not Reese Witherspoon. Reese Witherspoon did not star in 'Juno.' Therefore, the association between Reese Witherspoon and 'Juno' is incorrect. Movies About Tracey is likely 'Tracey Fragments (2007)', a movie in which Ellen Page also starred as the lead character, Tracey Berkowitz. Ellen Page (Elliot Page) connects both movies.

Final answer: Ellen Page (now Elliot Page)

Question: \{question text\} \\

Initial Answer: \{initial answer\} \\
Write a correction if the initial answer is incorrect.
\end{quote}

\end{document}